\pdfoutput=1

\documentclass[11pt]{article}

\usepackage{placeins}
\usepackage{graphicx}
\usepackage{soul}
\usepackage[normalem]{ulem}
\useunder{\uline}{\ul}{}
\usepackage{comment}
\usepackage[]{emnlp2021}

\usepackage{times}
\usepackage{latexsym}

\usepackage[T1]{fontenc}

\usepackage[utf8]{inputenc}

\usepackage{amsmath}
\usepackage{amssymb}
\usepackage{natbib}

\usepackage{microtype}

\usepackage{chngpage}

\usepackage{tikz}
\usetikzlibrary{snakes, arrows, shapes}

\usetikzlibrary{
	arrows.meta, 
	fit, 
	positioning, 
}

\tikzset{
	conn/.style={ 
		-{Straight Barb[angle=60:2pt 3]}, 
		thick, 
	},
	cell/.style={
		rectangle, draw, thick,
		minimum width=3.5em,
		minimum height=2.5em,
		line width=0.25mm,
		node distance=0.5ex and 2em
	},
	plain/.style={ 
		node distance = 0.5ex and 2em
	},
}

\title{Prediction of Listener Perception of Argumentative Speech in a Crowdsourced Dataset Using (Psycho-)Linguistic and Fluency Features}

\author{{\it Yu Qiao$^1$, Sourabh Zanwar$^1$, Rishab Bhattacharyya$^1$} \\ {\it Daniel Wiechmann$^2$, Wei Zhou$^{1,3}$, Elma Kerz$^1$ and Ralf Schlüter$^{1,3}$} \vspace{3mm} \\  $^1$RWTH-Aachen University, $^2$University of Amsterdam, $^3$AppTek GmbH \\ \texttt{\{yu.qiao,sourabh.zanwar,rishab.bhattacharyya\}@rwth-aachen.de},\\ \texttt{\{zhou, schlueter\}@cs.rwth-aachen.de,}\\ \texttt{d.wiechmann@uva.nl, elma.kerz@ifaar.rwth-aachen.de}}

\begin{document}
\maketitle
\begin{abstract}
One of the key communicative competencies is the ability to maintain fluency in monologic speech and the ability to produce sophisticated language to argue a position convincingly. In this paper we aim to predict TED talk-style affective ratings in a crowdsourced dataset of argumentative speech consisting of 7 hours of speech from 110 individuals. The speech samples were elicited through task prompts relating to three debating topics. The samples received a total of 2211 ratings from 737 human raters pertaining to 14 affective categories. We present an effective approach to the classification task of predicting these categories through fine-tuning a model pre-trained on a large dataset of TED talks public speeches. We use a combination of fluency features derived from a state-of-the-art automatic speech recognition system and a large set of human-interpretable linguistic features obtained from an automatic text analysis system. Classification accuracy was greater than 60\% for all 14 rating categories, with a peak performance of 72\% for the rating category `informative'. In a secondary experiment, we determined the relative importance of features from different groups using SP-LIME.

\end{abstract}

\section{Introduction}
Effective communication skills are a cornerstone of several cognitive and socio-emotional aspects of development and a key component of personal satisfaction, academic achievement, and career success \citep{morreale2008communication}. Given the importance of these skills, it is hardly surprising that they are an ongoing subject of study by researchers from numerous disciplines \citep{greene2003handbook, backlund2015communication}. In recent years, there has been an increasing effort to use natural language processing techniques in combination with machine learning to predict human evaluation of speech performance. Existing studies in this research area have already provided valuable insights into the correlates of such evaluations \citep{weninger2012voice,weninger2013words,tanveer2019predicting,kerz-etal-2021-language,reddy2021modeling}. However, these studies have been typically confined to predicting affective ratings of speeches produced by domain experts, such as in the context of TED Talks\footnote{TED (Technology, Entertainment and Design) Talks are designed to provide enlightening insights on various topics (https://www.ted.com/). TED presenters are often selected not only on the basis of their expertise on a given topic but also for their ability to effectively and succinctly communicate.} and are based on large-scale datasets that include over 2,000 talks with more than 500 hours of speaking time and more than 5.5 million ratings.

In this paper, we take important first steps toward extending this line of research to the prediction of human affective evaluations in smaller crowdsourced samples of argumentative speech by less experienced speakers. We build models to predict fourteen TED talk-style categories (beautiful, confusing, courageous, fascinating, funny, informative, ingenious, inspiring, jaw-dropping, longwinded, obnoxious, OK, persuasive, unconvincing) by fine-tuning models pre-trained on a large dataset of TED talks. The models incorporate a large number of human-interpretable (psycho-)linguistic features in combination with fluency features derived from state-of-the-art automatic text analysis and speech recognition systems. We also perform feature ablation experiments using SP-LIME to determine the relative importance of such features for the prediction of the affective categories.

\begin{figure*}
	\centering
	\includegraphics[width = 0.7\textwidth]{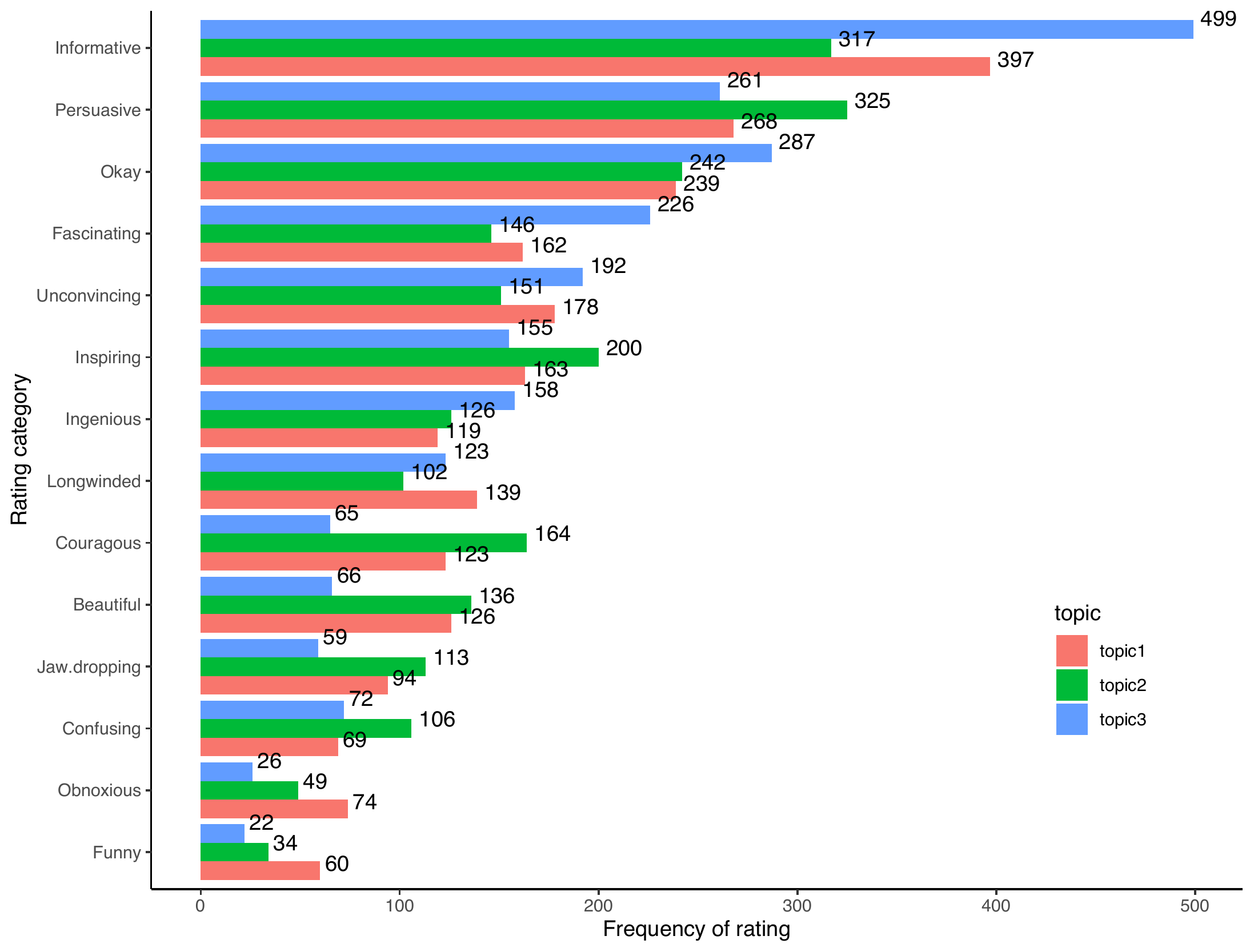}
	\caption{Frequency of TED-style affective rating by topic across the fourteen rating categories. Label frequencies vary strongly for different labels. For instance, the label `informative' has been assigned 1213 while ‘confusing’ is only assigned 116 times.}
	\label{fig:distributions}
\end{figure*}

The remainder of the paper is organized as follows: Section 2 presents a brief overview of related work. Section 3 describes the experimental setup: Section 3.1 introduces the two datasets alongside with affective ratings. Sections 3.2 and 3.3 describe the measurement of (psycho-)linguistic and fluency features. Section 3.4 gives a description of the classification model
architecture and the approach used to evaluate feature importance. Section 4 presents the main results and discusses them. Finally, Section 5 summarizes the main findings reported in the paper and suggests future research directions.

\section{Related work}

Several previous studies have examined the relationship between linguistic features and human affective ratings. \citet{weninger2012voice} analysed 143 online speeches hosted on YouTube to classify individuals as achievers, charismatic speakers, and team players with 72.5\% accuracy on unseen data. \citet{weninger2013words} predicted the affective ratings for
online TED talks using lexical features, where online viewers assigned 3 out of 14 predefined rating categories that resulted in the affective state invoked in them listening to the talks. Their models reached average recall rates of 74.9 for positive categories (jaw-dropping, funny, courageous, fascinating, inspiring, ingenious, beautiful, informative, persuasive) and 60.3 for neutral or negative ones (OK, confusing, unconvincing, long-winded, obnoxious). \citet{tanveer2019predicting} predicted TED talk ratings from using psycholinguistic language features, prosody and narrative trajectory features. Using three neural network architectures they were able to predict TED different ratings with an average AUC of 0.83. \citet{kerz-etal-2021-language} showed that a recurrent neural network classifier trained solely on in-text distributions of language features can achieve relatively high accuracies (>70\%) on eight of fourteen TED rating categories. Moreover, their ablation experiments showed that the best predictive feature sets belong to LIWC-style psycholinguistic features and n-gram frequency measures that capture the use of genre-specific multi-word sequences. 
While previous work has provided interesting insights into the relationship between textual features in speech samples and listeners' affective ratings, the question is to what extent similar patterns can be observed in smaller datasets of non-professional speakers.

\section{Experimental Setup}
\subsection{Dataset}
The dataset of argumentative speech used in this study consisted of 7 hours of speech from 111 individuals (aged 18-30 years) collected through the Amazon Mechanical Turk (AMT) crowdsourcing platform. The speech samples were elicited through prompts relating to three debating topics: (A) Climate change is the greatest threat facing humanity today, (B) People should be legally required to get vaccines, and (C) The development of Artificial Intelligence will help Humanity. The number of speech samples was evenly distributed across the three topics (N=37 speeches per topic). We were able to analyze only 110 out of 111 speeches, since one speech had to be excluded due to a strong distortion. The speech samples were rated by another group of 737 crowdworkers, who were asked to assign up to three of the 14 impression-related labels given to viewers of TED.com (beautiful, confusing, courageous, fascinating, funny, informative, ingenious, inspiring, jaw-dropping, longwinded, obnoxious, OK, persuasive, unconvincing). Each rater evaluated three speeches amounting to a total of 2211 affective ratings. Figure \ref{fig:distributions} shows the frequency distributions of the fourteen rating categories by topic. Each speech was rated by an average of 19.92 human raters. Table \ref{tab:lableFreq} presents the minimum, maximum and mean label frequencies assigned to the 110 speeches per label.  Inter-rater reliability was substantial across categories with Cohen's Kappa scores ranging between 0.65 and 0.95 (see Table \ref{tab:interrater}). All speeches were manually transcribed by two transcribers. 

\begin{table}[]
\caption{Minimum, maximum and mean label frequency per label assigned to the 110 speeches in the dataset. }
\centering
\begin{tabular}{llll}
\hline
             & \multicolumn{3}{l}{Tag frequency} \\
\hline
Label          & min      & max         & mean     \\
\hline
\multicolumn{4}{c}{Positive}                     \\
\hline
Beautiful   & 0 & 17      & 2.95 \\
Courageous   & 0 & 18      & 3.17 \\
Fascinating & 0 & 22      & 4.81 \\
Funny       & 0 & 22      & 1.05 \\
Ingenious   & 0 & 16      & 3.63 \\
Informative  & 0 & 24  &   10.9   \\
Inspiring    & 0 & 20      & 4.67 \\
Jaw-dropping & 0 & 16      & 2.40 \\
Persuasive  & 0 & 23      & 7.69 \\
\hline
\multicolumn{4}{c}{Neutral / Negative}           \\
\hline
Confusing      & 0 & 17      & 2.23 \\
Longwinded   & 0 & 24      & 3.28 \\
Obnoxious    & 0 & 21      & 1.34 \\
Okay         & 0 & 25      & 6.92 \\
Unconvincing & 0 & 20      & 4.69 \\
\hline
\end{tabular}
\label{tab:lableFreq}
\end{table}

\begin{table}[]
\caption{Inter-rater reliability (Cohen's Kappa) across all rating categories. Each speech was rated by eleven to twenty-seven annotators (avg. 19.92 raters/speech)}
\centering
\begin{tabular}{lc}
\hline
Category     & Inter-rater reliability \\
&(Cohen's Kappa)\\
\hline
Persuasive   & 0.65   \\
Courageous   & 0.80    \\
Inspiring    & 0.75    \\
Jaw-dropping & 0.85    \\
Fascinating  & 0.73    \\
Beautiful    & 0.81    \\
Informative  & 0.66    \\
Ingenious    & 0.79    \\
Funny        & 0.95    \\
Unconvincing & 0.77    \\
Okay         & 0.72    \\
Confusing    & 0.87    \\
Obnoxious    & 0.93    \\
Long-winded  & 0.84    \\
\hline
Overall      & 0.79   \\
\hline
\end{tabular}
\label{tab:interrater}
\end{table}

To pre-train our models we constructed a large auxiliary dataset of TedTalks gathered from the ted.com website\footnote{\url{https://www.ted.com/}}. We crawled the site and obtained every TED Talk transcript and its metadata from 2006 through 2017, which yielded a total of 2668 talks. Viewers on the Internet can vote for three impression-related labels out of the 14 types of listed above. The labels are not mutually exclusive and users can select up to three labels for each talk. If only a single label is chosen, it is counted three times. All talks that featured more than one speaker as well as talks that centered around music performances were removed. This resulted in a dataset of 2392 TED talks with a total number of views of 4139 million and a total number of 5.89 million ratings. All ratings were normalized per million views to account for differences in the amount of time that talks have been online. For both datasets, all ratings were binarized by their medians, such that each category has a value 1 when the rating of a text in this category was above or equal to the median and 0 if not.

\subsection{Measurement of (psycho-)linguistic features}

\begin{figure}
    \centering
    \includegraphics[width = 0.5\textwidth]{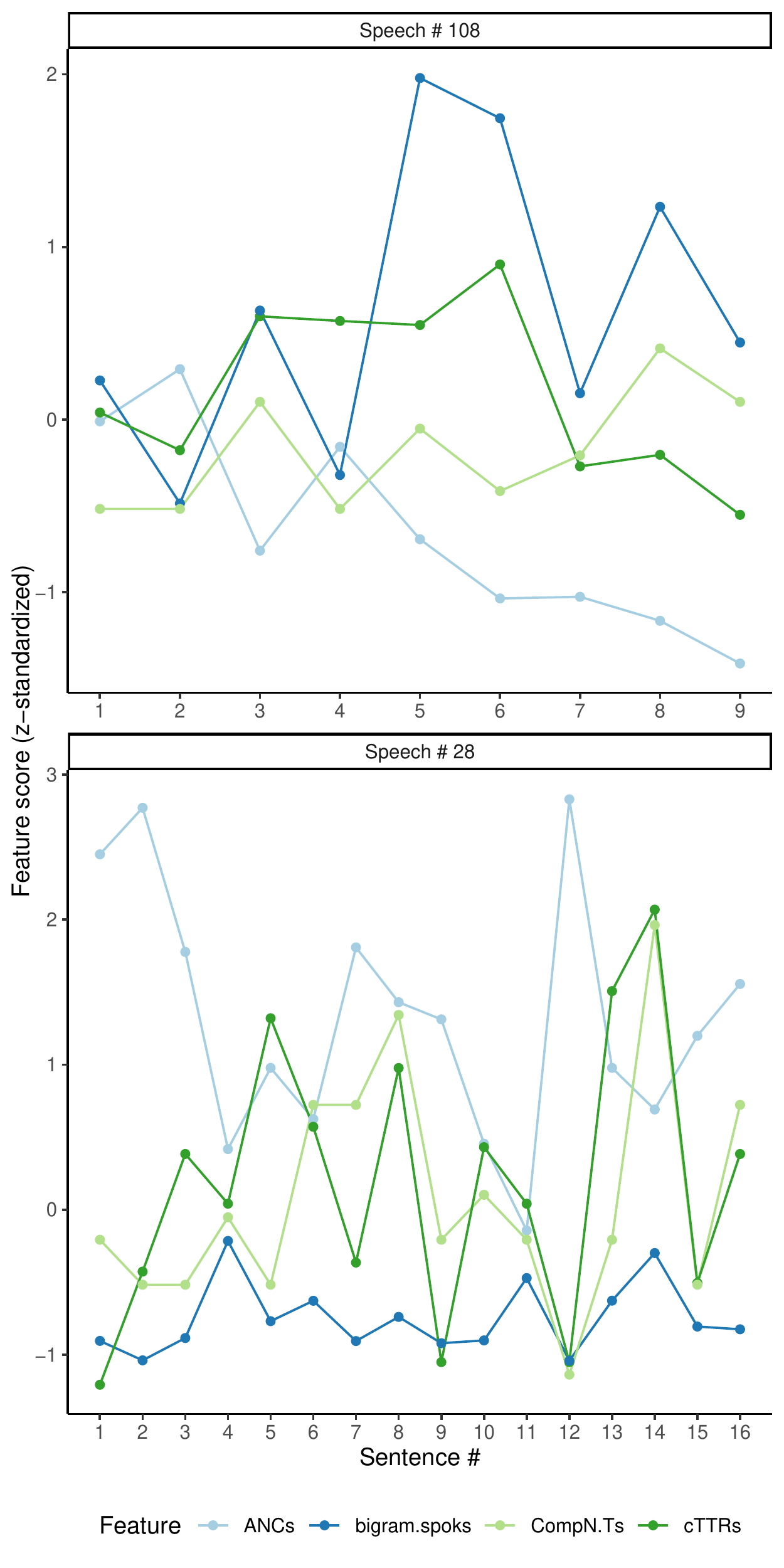}
    \caption{Examples of `complexity contours', i.e. sentence-level measurements of language features, for four selected measures (ANCs, bigram.spoks, CompN.Ts, cTTRs) of two randomly selected speeches (Speech \# 28 and Speech \# 108). For the purposes of this visualization, all features have been z-standardized. A score of 0 denotes the mean score of the entire group of participants. (ANCs = Lexical Sophistication (American National Corpus), bigram.spoks = weighted bigram frequency score (spoken register), CompN.Ts = Complex nominal per T-Unit, cTTRs = corrected Type-Token ratio)}
    \label{fig:examples}
\end{figure}

We extracted a total of 354 features that fall into six categories:  (1) measures of syntactic complexity, (2) measures of lexical richness, (3) register-based n-gram frequency measures, (4) information-theoretic measures, and (5) LIWC-style (Linguistic Inquiry and Word Count) measures, and (6) word prevalence measures. Sentence-level measurements of all features were obtained using CoCoGen, a computational tool that implements a sliding window technique to calculate so called `complexity contours' representing the within-text distributions of scores for a given language feature (for current applications of the tool in the context of text classification, see \cite{kerz2020becoming,qiao2020language,strobel2020relationship}). Tokenization, sentence splitting, part-of-speech tagging, lemmatization and syntactic PCFG parsing were performed using Stanford CoreNLP \cite{manning2014stanford}. Figure \ref{fig:examples} presents examples of the extracted complexity contours for four  selected  measures of two randomly selected speeches.  As is evident in the graphs, all features scores fluctuate within each speech and often display `compensatory' behavior, such that high scores in one feature are accompanied by low scores on another. 

\subsection{Extraction of fluency features}

To derive fluency features, the pretrained hybrid Hidden Markov Model-based automatic speech recognition (ASR) system from \cite{zhou2020rwth} was used, which showed state-of-the-art performance on the 2nd release of TED-LIUM task (TLv2) \cite{rousseau2014enhancing}. The same LSTM-based language models (LM) as in \cite{zhou2020rwth} were used for recognition, which were trained on the TLv2 LM training data. The bidirectional long short-term memory (BLSTM) based acoustic model (AM) was fine-tuned on the present dataset. The 7 hours of acoustic training data were divided into training, dev(elopment), testing sets of 4 hours, 1 hour and 2 hours, respectively. 
The hyperparameters for fine tuning were optimized on the dev set, which yielded a constant learning rate of $2\times 10^{-5}$, LM scale of 10.0 and a LM look ahead factor of 0.9. Speaker adaptation techniques are commonly applied to account for speaker variability and to improve ASR performance. Following \citet{zhou2020rwth}, here we adopted the i-vectors-based speaker embedding approach. To further improve the performance of our ASR system, confusion network decoding was applied. The final fine-tuned ASR system achieved a WER of 18.4\% on the test set. We derived three fluency features from the ASR system that fall into three classes. (1) \textit{Silent pauses} - Durations of pauses were calculated from forced alignment. A silent pause threshold of 250 msec was used for pause counts. In addition, we calculated the total pause duration per sentence (in sec). (2) \textit{Speed of articulation} We enriched the output of the ASR with syllable counts from the Carnegie Mellon University Pronouncing Dictionary\footnote{http://www.speech.cs.cmu.edu/cgi-bin/cmudict}. Next to measuring articulation rate in terms of words per minute, we derived mean syllable durations an well as syllables per minute for each utterance in the speech data. (3) \textit{Filled pauses} - Next to the number and total duration of silent pauses, we derived frequency total and normalized counts of filled pauses, i.e. hesitation markers identified by human transcribers.

\subsection{Modeling Approach}

	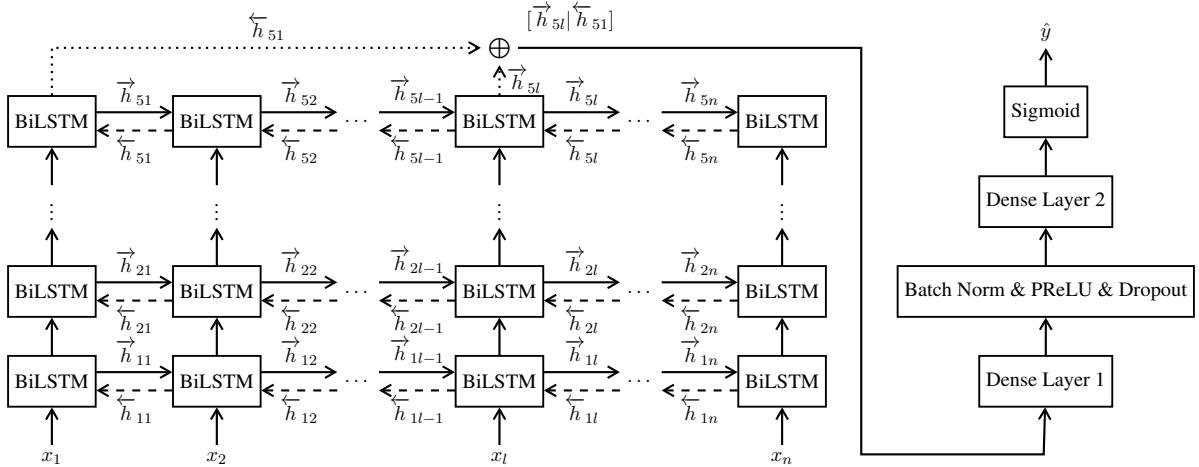
\begin{figure*}
		\centering
		\begin{tikzpicture}[scale=0.7, every node/.style={scale=0.7}]
		\node[cell] (BiLSTM00) {BiLSTM};
\node[cell, above=0.5cm of BiLSTM00] (BiLSTM10) {BiLSTM};
\foreach \x [remember=\x as \lastx (initially 0)] in {1,...,1} {
	\node[cell, right=1cm of BiLSTM0\lastx] (BiLSTM0\x) {BiLSTM};
	\node[cell, right=1cm of BiLSTM1\lastx] (BiLSTM1\x) {BiLSTM};
}

\node[right=1cm of BiLSTM01] (BiLSTM02) {$\dots$};
\node[right=1cm of BiLSTM11] (BiLSTM12) {$\dots$};

\node[cell, right=1cm of BiLSTM02)] (BiLSTM03) {BiLSTM};
\node[cell, right=1cm of BiLSTM12] (BiLSTM13) {BiLSTM};

\node[right=1cm of BiLSTM03] (BiLSTM04) {$\dots$};
\node[right=1cm of BiLSTM13] (BiLSTM14) {$\dots$};

\node[cell, right=1cm of BiLSTM04)] (BiLSTM05) {BiLSTM};
\node[cell, right=1cm of BiLSTM14] (BiLSTM15) {BiLSTM};

\node[below=.5cm of BiLSTM00] (x1) {$x_1$};
\node[below=.5cm of BiLSTM01] (x2) {$x_2$};
\node[below=.5cm of BiLSTM03] (xl) {$x_l$};
\node[below=.5cm of BiLSTM05] (xn) {$x_n$};
\draw[conn] (x1.north) -- (BiLSTM00.south);
\draw[conn] (x2.north) -- (BiLSTM01.south);
\draw[conn] (xl.north) -- (BiLSTM03.south);
\draw[conn] (xn.north) -- (BiLSTM05.south);

\foreach \x in {0,1,3,5} {
	\draw[conn] (BiLSTM0\x.north) -- (BiLSTM1\x.south);
}

\foreach \x in {0,1,3,5} {
	\node[above=.5cm of BiLSTM1\x] (BiLSTM2\x) {$\vdots$};
	\draw[conn] (BiLSTM1\x.north) -- (BiLSTM2\x.south);
}

\foreach \x in {0,1,3,5} {
	\node[cell, above=.5cm of BiLSTM2\x] (BiLSTM3\x) {BiLSTM};
	\draw[conn] (BiLSTM2\x.north) -- (BiLSTM3\x.south);
}

\node[right=1cm of BiLSTM31] (BiLSTM32) {$\dots$};
\node[right=1cm of BiLSTM33] (BiLSTM34) {$\dots$};

\foreach \y/\j in {0/1,1/2,3/5} {
	\foreach \x/\z [remember=\x as \lastx (initially 0)] in {1/1, 2/2,3/l-1, 4/l, 5/n} {
		\draw[conn] ([yshift=1ex]BiLSTM\y\lastx.east) -- ([yshift=1ex]BiLSTM\y\x.west) node[above,midway] {$\overrightarrow{h}_{\j\z}$};
		\draw[conn, dashed] ([yshift=-1ex]BiLSTM\y\x.west) -- ([yshift=-1ex]BiLSTM\y\lastx.east) node[below,midway] {$\overleftarrow{h}_{\j\z}$};
	}
}

\node[above=0.4cm of BiLSTM33] (ADD) {$\bigoplus $};
\node[cell, right=2cm of BiLSTM05] (fc1) {Dense Layer 1};
\node[cell, above=.5cm of fc1] (relu) {Batch Norm \& PReLU \& Dropout};
\node[cell, above=.5cm of relu] (fc2) {Dense Layer 2};
\node[cell, above=.5cm of fc2] (sig) {Sigmoid};
\node[above=.5cm of sig] (y) {$\hat{y}$};

\draw[conn] (ADD)--++(6.8,0)--++(0, -7.7)--++(3.43, 0)--(fc1);
\foreach \x [remember=\x as \lastx (initially fc1)] in {relu, fc2, sig, y} {
	\draw[conn] (\lastx) -- (\x);
}
\draw[conn, dotted] (BiLSTM30) |- (ADD) node[above, 	pos=0.75] {$\overleftarrow{h}_{51}$};
\draw[conn, dotted] (BiLSTM33) -- (ADD) node[right, midway] {$\overrightarrow{h}_{5l}$};
\node[above right=-.1cm and 0.01 of ADD] {$[\overrightarrow{h}_{5l} |\overleftarrow{h}_{51} ]$};
		\end{tikzpicture}
		\caption{Roll-out of the RNN model based on complexity contours} \
	\label{fig:model}
	\vspace{-6mm}
\end{figure*}

We trained fourteen recurrent neural network (RNN) classifiers -- one for each affective category -- consisting of five bidirectional long short-term memory (LSTM) layers with a hidden state dimension of 400 (see Figure \ref{fig:model}).\footnote{We also performed experiments with a multiclass-classification model. However, we found that training all 14 categories together had detrimental effects on the classification accuracy of the more predictive categories.} The input to model is a sequence  $X=(x_1, x_2,\dots, x_l, x_{l+1}, \dots, x_n)$, where $x_i$, the output of CoCoGen for the $i$th window of a document, is a 354 dimensional vector, $l$ is the length of the sequence, $n\in \mathbb{Z}$ is a number, which is greater or equal to the length of the longest sequence in the dataset and $x_{l+1},\cdots, x_n$ are padded $\mathbf{0}$-vectors. To predict the class of a sequence, we concatenate the hidden variable of the last LSTM cell in layer 5 $\overrightarrow{h}_{5l}$, i.e. the hidden variable of 5$th$ RNN layer right after the feeding of $x_l$, with the hidden variable of the last LSTM cell in the backward direction $\overrightarrow{h}_{51}$. The result vector of concatenation $[\overrightarrow{h}_{5l}|\overrightarrow{h}_{51}]$ is then transformed through a feed-forward neural network. The feed-forward neural-network consists of two fully connected layers (dense layer), whose output dimensions are 400, 1. Between the first and second fully connected layer, a Batch Normalization layer, a Parametric Rectifier Linear Unit (PReLU) layer and a dropout layer were added. Before the final output, a sigmoid layer was applied. As the loss funtion, binary cross entropy loss was used. Our implementation uses the PyTorch library \citet{paszke2017pytorch}. Input data were standardized within the training folds. Supervised pre-training on a large external dataset followed by domain-specific fine-tuning on a small dataset was presented by \citet{girshick2014rich} as an effective approach for modeling scarce training data. Here, we follow that approach by first training a BLSTM classifier on the TED dataset used in \citet{kerz-etal-2021-language} and then fine-tuning the obtained model on the present dataset for each affective rating category.  To include both fluency features and the topic of the speech (encoded as one-hot vectors), we replaced the last FC layer with two consecutive randomly initialized FC layers whose input dimension is that of the removed FC, extended by the total number of dimensions of the fluency features (7) and the topic vector (3). The output dimension is 1. The new FC layers are activated by PReLU. To suppress overfitting, a dropout layer with a dropout rate of 0.5 was added between the two new FC layers. For fine-tuning, the BLSTM output was concatenated with the fluency features and topic vectors and fed into the newly added FC layers. We used a smaller stack size of 8 and a smaller learning rate of 0.0001. To mitigate the effects of class distribution imbalance (ratio be-tween minority and majority class smaller than 4:6) observed for some of the evaluation categories (\textit{funny}, \textit{obnoxious}, \textit{confusing}, \textit{jaw.dropping}), we assume a class weight of $1-p(c)$, where $c\in\{0,1\}$ is the class label of the false and true class, respectively, and $p(c)$ is the empirical probability of class $c$. All fourteen models were evaluated through 5-fold cross validation using an 80/20 training/testing split. All hyperparameters were optimized using grid search.

For the feature ablation, we employed Submodular Pick Lime (SP-LIME; \citet{ribeiro2016should}), a method to construct a global explanation of a model by aggregating the weights of the linear models. The linear models serve as approximations of a complex model around small regions on the data manifold. To this end we first constructed local explanations using LIME. Analogous to super-pixels for images, we categorized our features into seven groups – six (psycho-)linguistic groups plus fluency – and used binary vectors $z\in\{0,1\}^{d}$ to denote the absence and presence of feature groups in the perturbed data samples, where $d$ is the number of feature groups. Here, absent means that all values of the features in the feature group are set to 0, and present means that their values are retained. For simplicity, a linear regression model was chosen as the local explanatory model. An exponential kernel function with Hamming distance and kernel width $\sigma=0.75\sqrt{d}$ was used to assign different weights to each perturbed data sample. After constructing their local explanation for each data sample in the original dataset, the matrix $W\in\mathbb{R}^{n\times d}$ was obtained, where $n$ is the number of data samples in the original dataset and $W_{ij}$ is the $j$th coefficient of the fitted linear regression model to explain data sample $x_i$. The global importance score of the SP-LIME for feature $j$ can then be derived by: $I_j = \sqrt{\sum_{i=1}^n |W_{ij}|}$

\section{Results and Discussion}

\begin{table}
\setlength\tabcolsep{8pt}
    \centering
    \caption{Performance of the fine-tuned models. The `*' indicates an imbalanced class distribution (ratio between minority and majority class smaller than 4:6). `Total Avg' = average including imbalanced categories; `Avg' = imbalanced categories not included}
   \begin{tabular}{lcccc}
\hline
      Category &   Acc &  Rec &  Prec &    F1 \\
\hline
        Funny* &  0.88 &    0.21 &       0.60 &  0.32 \\
    Obnoxious* &  0.82 &    0.16 &       0.43 &  0.23 \\
   Informative &  0.72 &    0.78 &       0.69 &  0.74 \\
    Courageous &  0.71 &    0.71 &       0.71 &  0.71 \\
    Confusing* &  0.71 &    0.34 &       0.57 &  0.43 \\
 Jaw.dropping* &  0.67 &    0.45 &       0.56 &  0.50 \\
     Beautiful &  0.66 &    0.71 &       0.60 &  0.65 \\
    Longwinded &  0.66 &    0.43 &       0.61 &  0.51 \\
          Okay &  0.65 &    0.67 &       0.62 &  0.64 \\
   Fascinating &  0.64 &    0.55 &       0.67 &  0.60 \\
     Inspiring &  0.63 &    0.63 &       0.60 &  0.62 \\
  Unconvincing &  0.61 &    0.47 &       0.60 &  0.53 \\
    Persuasive &  0.61 &    0.57 &       0.61 &  0.59 \\
     Ingenious &  0.60 &    0.65 &       0.59 &  0.62 \\
     \hline
     Total Avg &  0.68 &    0.52 &       0.60 &  0.55 \\
           Avg &  0.65 &    0.62 &       0.63 &  0.62 \\
\hline
\end{tabular}
    \label{tab:model_performance}
    \vspace{-5mm}
    \end{table}


 The performance metrics of the fourteen fine-tuned BLSTM classification models (global accuracy, precision, recall, and F1 scores, all macro averages) are shown in Table \ref{tab:model_performance}. The highest accuracy was obtained for the \textit{funny} category (88.2\%) and the lowest for the \textit{ingenious} category (60.0\%). On average, a classification accuracy of 68.4\% was achieved. However, due to the unbalanced class distributions of the categories, the accuracy results of highly unbalanced categories (marked with `*' in Table \ref{tab:model_performance}) should be interpreted with caution. For these categories, despite applying class weights and assigning a higher penalty to minority class samples to mitigate the effects of class imbalance, the classification results were still strongly influenced by the empirical distributions of class labels, as indicated by their low F1 scores. When all highly imbalanced categories are removed, our classification models achieve an average accuracy of 64.9\%.  Importantly, in the current study peak performance (72\% accuracy) was achieved for the category \textit{informative}, indicating that the (psycho-)linguistics and fluency-related features considered in this study are important predictors for this aspect of argumentative speech. Comparison of the results with those obtained by  \citet{kerz-etal-2021-language} revealed that in both cases the categories \textit{courageous} and \textit{beautiful} are among the best predicted categories while \textit{unconvincing} and \textit{ingenious} were the least predictable ones. Noticeable differences in classification accuracy were observed for categories \textit{persuasive}, \textit{longwinded} and \textit{informative}.   In \citet{kerz-etal-2021-language}, the persuasive category was among the best predicted (ranked 1 of 14), while in the current study it achieved comparatively low classification accuracy (ranked 13 of 14); in contrast, the informative category achieved rank 6 of 14 in \citet{kerz-etal-2021-language}, while in the current study it achieved rank 3 of 14 accuracy; the longwinded category appeared at the lowest rank in \citet{kerz-etal-2021-language}, while here it achieved rank 8 of 14 accuracy.

\begin{table}[]
    \centering
    \caption{Result of feature ablation: FI = feature importance score from SP-LIME; `Syntax' = syntactic complexity group; `Inf. Th' = Information theoretic complexity group}
    \begin{tabular}{lrlrlr}
\hline
\multicolumn{6}{c}{Rating category}\\
\hline
    \multicolumn{2}{c}{Informative} & \multicolumn{2}{c}{Persuasive} & \multicolumn{2}{c}{Unconvincing} \\
          Group & FI &           Group & FI &           Group & FI \\
\hline
          N-gram &  2.67 &           N-gram &  5.83 &           N-gram &  4.75 \\
           LIWC &  2.31 &            LIWC &  4.40 &            LIWC &  3.24 \\
      Syntax &  1.89 &      Preval. &  3.73 &      Preval. &  2.77 \\
     Preval. &  1.67 &       Syntax &  2.92 &       Syntax &  2.54 \\
        Lexical &  1.39 &  Lexical &  2.90 &  Lexical &  2.01 \\
     Inf. Th &  0.61 &      Inf. Th &  1.32 &      Inf. Th &  1.06 \\
        Fluency &  0.51 &         Fluency &  0.39 &         Fluency &  0.39 \\
\hline
\end{tabular}
    \label{tab:ablation}
    \vspace{-4mm}
\end{table}

\begin{figure*}
    \centering
    \includegraphics[width = 1\textwidth]{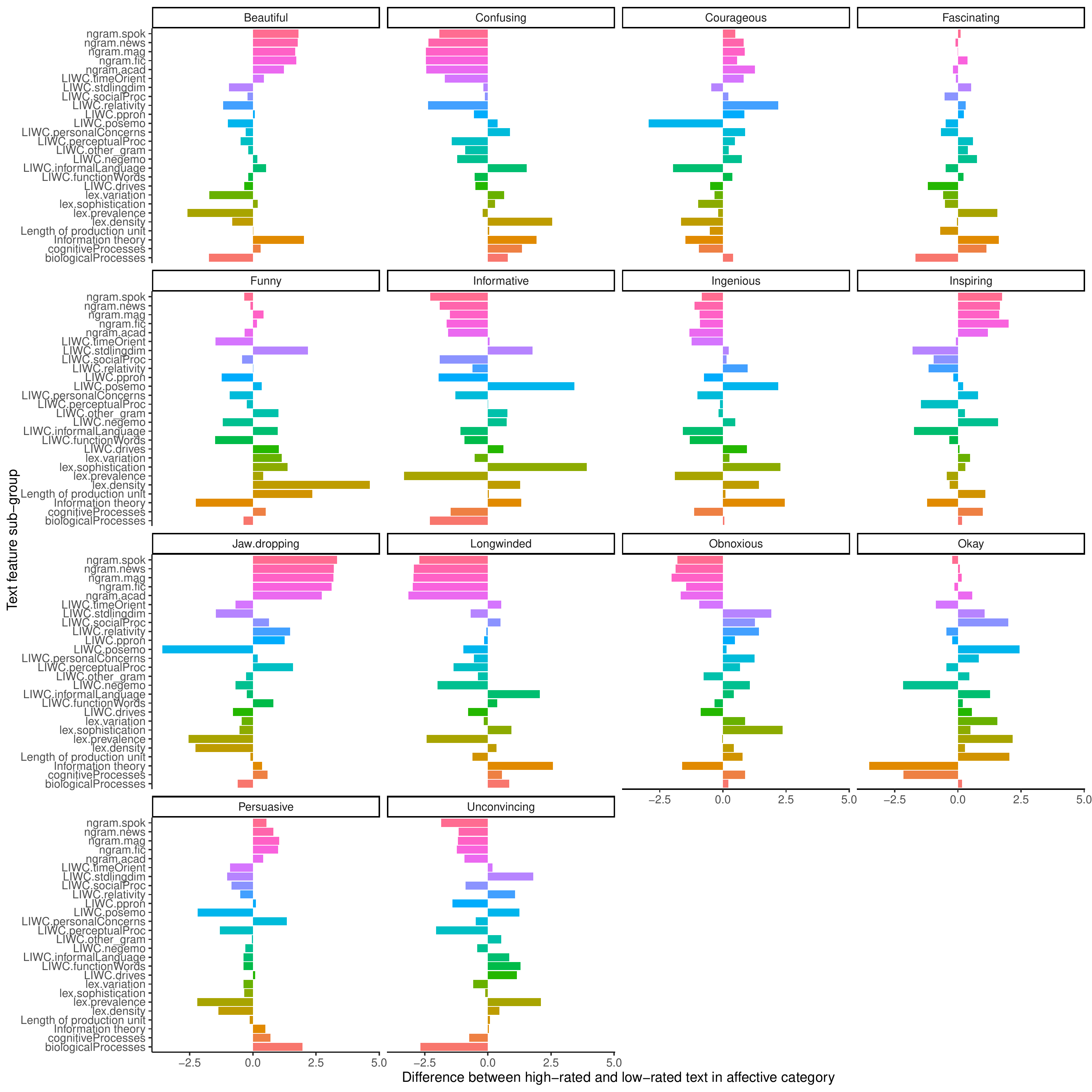}
    \caption{Differences between the group means of higher-rated speeches and lower-rated speeches by feature subgroup across the 14 rating categories. Bars extending to the right indicate that higher-rated speeches display higher scores on a given feature subgroup. Bars extending to the left indicate that higher-rated speeches display lower scores on that feature subgroup.}
    \label{fig:detailed}
\end{figure*}

 Table \ref{tab:ablation} shows the result of feature ablation results of 3 selected categories (\textit{informative, persuasive and unconvincing}). These categories were selected as they are (a) relevant given the communicative goals of argumentative speech, (b) their class label distributions were relative balanced and (c) relatively high prediction accuracy were obtained for these categories. These results reveal that – as was observed in \citet{kerz-etal-2021-language} –  the classification accuracy was mainly driven by LIWC-style and N-gram based features across categories, indicating that prediction of affective ratings is highly impacted by the same linguistic features in both expert and novice speakers. A closer examination of how individual features and sub-groups within the feature-groups distinguished between higher-rated and lower-rated speeches in a given rating category revealed some interesting patterns. To disclose such patterns, we dichotomized rating scores from each rating category  using median splits and determined the differences in feature scores between the group means of higher-rated speeches and lower-rated speeches (M\textsubscript{high\ rated}  - M\textsubscript{low\ rated}). For reasons of exposition, we focus on the results for some selected categories. A visualization of the results for all 14 rating categories is presented in Figure \ref{fig:detailed}. For example, higher-rated speeches in the \textit{informative} rating category are characterized by high scores on language features pertaining to lexical sophistication, indicating that these talks use words that are advanced and infrequent. They also comprise a high proportion of words relating to positive emotions. At the same time, these speeches are characterized by lower scores on ngram-frequency measures across all five language registers (spoken, fiction, news, magazine and academic language). In contrast, higher-rated speeches in the \textit{courageous} rating category are associated with higher scores on ngram-frequency measures and LIWC-style relativity words that concern time but score very low on words relating to positive emotions. Like informative speeches, confusing speeches score very low on the ngram-frequency measures but their lexical sophistication is much lower. These speeches are further characterized by larger proportions of words associated with informal language use.

\section{Conclusions}

The ability to communicate competently and efficiently yields innumerable benefits across a range of social arenas, including the enjoyment of congenial personal relationships, educational success, career advancement and, more generally, successful participation in the complex communicative environments of the 21st century. This paper contributes to the growing body of research that relies on automatic speech evaluation and machine learning to better understand what makes a speech effective. Specifically, we demonstrate an effective approach to predicting human ratings of small samples of argumentative speeches produced by less experienced speakers by fine-tuning a model pre-trained on a large dataset of public TED Talks speeches. Using a combination of fluent features derived from a fine-tuned automatic speech recognition model, combined with a large set of human-interpretable linguistic features obtained from an automatic text analysis system, we were able to achieve a prediction accuracy of 72\% for the \textit{informative} evaluation category and an average of 68.4\% across all fourteen categories studied. The results of the study show that the proposed approach provides a viable methodological basis for future research on human perception of speech based on crowdsourced datasets.
In future work, we intend to extend the set of fluency features used here to include other relevant features related to additional sub-dimensions of perceived fluency proposed in the literature, such as repair or prosodic measures. We further plan to investigate the extent to which the relationships between linguistic features and affective ratings are mediated by sociodemographic and personality characteristics of both speakers and listeners. Finally, we plan to conduct experiments comparing the effects of using automatically generated speech transcripts versus human-generated transcripts on the subsequent measurement of text features.
\newpage

\bibliography{anthology,eacl2021}
\bibliographystyle{acl_natbib}

\end{document}